\pdfoutput=1 % tell arXiv to use pdflatex (PDF/PNG figures)

\documentclass[runningheads]{llncs}
\usepackage[T1]{fontenc}
\usepackage[misc]{ifsym} % provides \Letter (envelope) for corresponding author
\usepackage{bbm}
\usepackage{multirow}

\usepackage{graphicx}
\usepackage{booktabs}
\usepackage{amsmath}
\usepackage{amsfonts}
\usepackage[hidelinks]{hyperref} % clickable references/URLs for the arXiv version

\begin{document}
\title{Improving Calibration of Black-Box Radiology \\ AI Using Test-Time Augmentation}
\titlerunning{TTA for Black-Box Radiology AI Calibration}
\author{Nathan Le\inst{1}$^{\star}$ \and
    Magdalini Paschali\inst{1}$^{\star}$ \and
    Arogya Koirala\inst{1} \and
    Andrew Johnston\inst{1} \and
    Zhongnan Fang\inst{1} \and
    David B. Larson\inst{1} \and
    Akshay S. Chaudhari\inst{1,2} \and
    Camila~Gonzalez\inst{1,3}\textsuperscript{(\Letter)}}

\authorrunning{N. Le, M. Paschali et al.}

\institute{Department of Radiology, Stanford University, Stanford, CA, USA \and
    Department of Biomedical Data Science, Stanford University, Stanford, CA, USA \and
    Department of Anesthesia, Intensive Care Medicine, and Pain Medicine, Medical University of Vienna, Spitalgasse 23, 1090 Vienna, Austria \\
    \email{camila.gonzalez@meduniwien.ac.at} \\
    \email{$^{\star}$ These authors contributed equally.}}
  
\maketitle              % typeset the header of the contribution
% Accepted-manuscript notice for the arXiv version (marker-less footnote, so it does
% not clash with the equal-contribution star in the author list).
{\renewcommand{\thefootnote}{}%
 \footnotetext{Accepted at the MICCAI 2026 Workshop on Uncertainty for Safe Utilization of
  Machine Learning in Medical Imaging (UNSURE 2026). This is the authors' accepted
  manuscript; it is not the Version of Record and does not reflect post-acceptance
  improvements or corrections. The Version of Record will be published by Springer in
  the Lecture Notes in Computer Science workshop proceedings.}}
\begin{abstract}
Radiology AI systems increasingly inform clinical decisions such as triage, follow-up imaging, and treatment planning. For these decisions to be made safely, model outputs must be well calibrated, meaning predicted probabilities accurately reflect true risk. Many standard techniques for improving calibration, such as MC Dropout and Deep Ensembles, require access to model parameters or retraining. However, proprietary clinical AI systems operate as black boxes, preventing access to the model's internals. To that end, we propose a model-agnostic framework for improving calibration of black-box models using clinically grounded test-time augmentation (TTA). Our framework applies geometric and physics-inspired 3D CT perturbations and learns probability-level aggregation strategies without access to model internals or the original training data. Across pulmonary embolism and intracranial hemorrhage detection tasks, \emph{DualTTA} achieved the strongest overall calibration among TTA methods, reducing the Expected Calibration Error by 54\% ($0.239 \rightarrow 0.109$) and 43\% ($0.051 \rightarrow 0.029$), respectively, while requiring only input-output access. Additionally, \emph{DualTTA} outperformed  uncertainty estimation techniques that require access to model internals, such as Temperature Scaling, MC Dropout, and Deep Ensembles, in most calibration metrics. These results demonstrate that learned TTA aggregation can improve the calibration of clinical AI systems, providing a practical approach for improving the reliability of black-box medical AI.

% Authors must provide keywords and are not allowed to remove this Keyword section.

\end{abstract}
\section{Introduction}

AI systems for radiology are deployed for triage, decision support, and worklist prioritization \cite{daye2022implementation}. In these workflows, model outputs influence clinical decisions, making calibration---the alignment between predicted probabilities and outcome likelihood---critical for safe use. Most commercial systems, however, operate as black boxes: only input–output behavior is observable, while model architecture, training data, and parameters remain inaccessible. This reality makes assessing reliability under real-world variability challenging.

Common sources of variability include acquisition differences and scanner- or institution-specific shifts, which can lead to performance gaps between pre-market evaluation and clinical use~\cite{huang2023inspect,zech2018variable}. Recent radiology reviews emphasize that reliable confidence estimation, including calibration and uncertainty quantification, is central for counteracting this risk \cite{mccrindle2021uncertaintyreview}, and automated monitoring approaches have been proposed to assess the reliability of deployed detection models in real time \cite{fang2025automated}.
However, commonly used uncertainty and calibration techniques, such as Monte Carlo Dropout \cite{gal2016dropout}  and Deep Ensembles \cite{lakshminarayanan2017simple}, require access to model parameters or retraining with the original training data. In practice, users of commercial AI tools are limited to the probability scores or, at times, only binary outputs that the closed system provides \cite{park2024caveats}, making these techniques infeasible for proprietary and black-box systems.

\begin{figure}[t]
    \centering
\includegraphics[width=\linewidth]{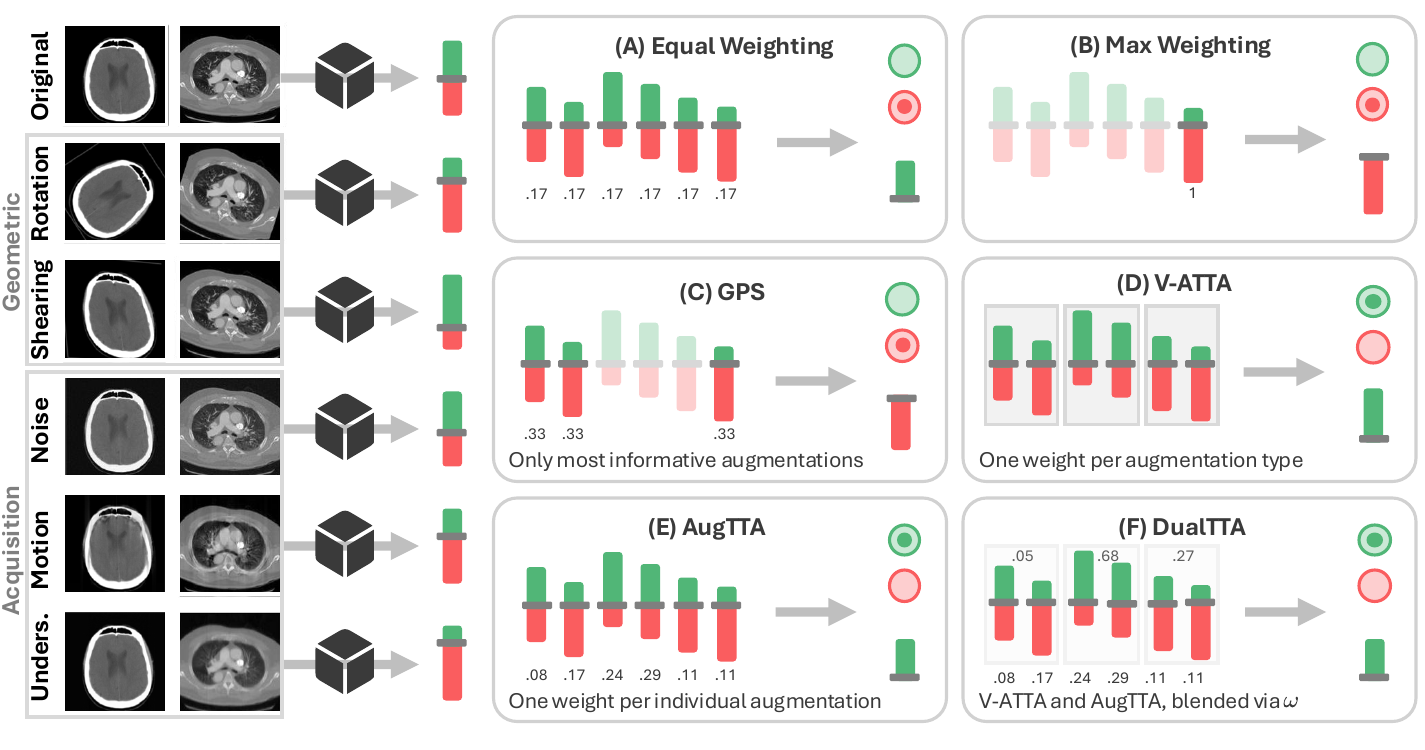}
    \caption{
        Proposed framework for improving black-box calibration. Left: a CT volume is perturbed by geometric and acquisition-based augmentations, and each perturbed volume is scored by the same frozen black-box model (identical black boxes), yielding one class probability estimate per augmentation (green/red bars). (A--F) The compared aggregation strategies combine these probabilities into a final, better-calibrated prediction (right side of each panel). Numbers denote aggregation weights, faded bars denote discarded predictions, and gray boxes group augmentations of the same type. In (F), \emph{DualTTA} additionally blends the aggregated and baseline outputs via a learned coefficient $\omega$.
    }
\label{fig:tta_overview}
\end{figure}

Test-time augmentation (TTA) provides a model-agnostic approach for analyzing black-box systems. By aggregating predictions across transformed inputs, TTA has been shown to improve robustness \cite{cohen2019certified,ma2022test} and to estimate uncertainty from the variation in predictions \cite{ayhan2020testtime,wang2019aleatoric}. Because it operates on the input–output interface, it is well-suited for proprietary models where internal access is unavailable. Prior work has used TTA with physics-based perturbations to probe the robustness of CT systems~\cite{highton2024robustness}, while separate work has developed aggregation strategies to improve robustness and uncertainty estimation \cite{shanmugam2021better,sherkatghanad2024baytta}.

However, prior work has employed TTA to improve robustness or to estimate uncertainty; to the best of our knowledge, TTA has not previously been proposed as a framework for the task of \emph{calibrating} black-box models, nor has the combination of clinically grounded 3D CT augmentations with optimized TTA aggregation been systematically evaluated for this purpose.

Our framework, shown in Fig.~\ref{fig:tta_overview}, provides a systematic
and externally applicable protocol for analyzing the confidence behavior of
proprietary radiology AI systems under realistic variability. Our
contributions are:
\begin{itemize}
    \item A model-agnostic framework that uses TTA for the novel task of
    improving the calibration of black-box classifiers, requiring only
    input--output access to the model and no training data;
    \item A clinically reviewed library of geometric and physics-inspired
    3D CT augmentations spanning three severity levels, evaluated with a
    range of probability-level aggregation strategies;
    \item \emph{DualTTA}, a learned aggregation strategy that unifies
    per-augmentation weighting~\cite{shanmugam2021better} with
    confidence-gated blending of baseline and aggregated
    predictions~\cite{conde2023approaching}. The two mechanisms address
    orthogonal failure modes: individual weights capture the unequal
    reliability of augmentations, while blending anchors the output to the
    trusted baseline prediction.
\end{itemize}
Through evaluation on pulmonary embolism and intracranial hemorrhage
detection, we show how different augmentation types, severity levels, and
aggregation strategies improve calibration, and find that learning separate
weights per augmentation effectively down-weights unstable predictions.
Our augmentation library, aggregation strategies, and evaluation code are
publicly available at \url{https://github.com/stanfordaide/TTA_Calibration}.

\section{Methods}

\paragraph{Problem Statement.} Let $f : \mathcal{X} \rightarrow \Delta^{C-1}$ denote a frozen black-box classifier mapping CT volumes $x \in \mathcal{X}$ to class probabilities. We assume no access to training data or model parameters. We define a library of $T$ augmentation types (e.g., rotation, noise), each applied at three severity levels, yielding $M = 3T$ augmentations $\{a_1, \ldots, a_M\}$, and obtain probability vectors $p_m(x) \in \Delta^{C-1}$ for $m = 0, \ldots, M$, where $m = 0$ denotes the unaugmented input. Each TTA aggregation method combines $\{p_m(x)\}_{m=0}^{M}$ into a calibrated output $\tilde{p}(x)$. 

\paragraph{Augmentation Library.}
All augmentations operate on 3D CT volumes at three severity levels (low, medium, and high). We include geometric perturbations (scaling, rotation, shearing) and acquisition-related perturbations (motion blur, noise, undersampling). All severity levels were reviewed by a board-eligible radiologist to ensure clinical plausibility; for example, medium undersampling approximates 50\% dose reduction.

\subsection{Aggregation Methods}

We compare the following strategies for combining the baseline and augmented predictions $\{p_m(x)\}_{m=0}^{M}$ into a single output $\tilde{p}(x)$, ranging from simple fixed rules to learned weighting schemes.

\smallskip
\noindent\textbf{No TTA} uses only the unaugmented output: $\tilde{p}_{\text{base}}(x) = p_0(x)$.

\smallskip
\noindent\textbf{Equal Weighting} averages probabilities across all $M+1$ predictions:
\begin{equation}
\tilde{p}_{\text{equal}}(x) = \frac{1}{M+1}\sum_{m=0}^{M} p_m(x).
\end{equation}

\smallskip
\noindent\textbf{Max Weighting}~\cite{hendrycks2016baseline} takes the element-wise maximum across augmentations for each class, favoring the most confident prediction across augmented views:
\begin{equation}
\tilde{p}_{\text{max}}(x)_c = \max_{m=0,\ldots,M} p_m(x)_c, \quad c = 1, \ldots, C.
\end{equation}

\smallskip
\noindent\textbf{GPS (Greedy Policy Search)}~\cite{lyzhov2020greedy} greedily selects up to 3 augmentations (including the original) that minimize validation Negative Log-Likelihood (NLL), then averages their predictions at test time:
\begin{equation}
\tilde{p}_{\text{GPS}}(x) = \frac{1}{|S|}\sum_{m \in S} p_m(x),
\end{equation}
where $S$ is the selected augmentation set.

\smallskip
\noindent\textbf{V-ATTA}~\cite{conde2023approaching} groups augmentations by type and learns one weight $\theta_t$ per type, shared across its three severity levels. Let $A_t \subset \{1, \ldots, M\}$ be the indices of type $t$ ($|A_t| = 3$), with type-averaged probability $\bar{p}_t(x) = \frac{1}{|A_t|}\sum_{m \in A_t} p_m(x)$. The aggregated prediction is blended with the baseline using a case-dependent coefficient $\omega(x)$, where $\text{normalize}(\mathbf{z}) = \mathbf{z}/\|\mathbf{z}\|_1$ denotes L1 normalization:
\begin{equation}
\hat{p}(x) = \text{normalize}\!\left(\sum_{t=1}^{T} \theta_t \, \bar{p}_t(x)\right),
\end{equation}
\begin{equation}
\tilde{p}_{\text{V-ATTA}}(x) = \bigl(1 - \omega(x)\bigr)\, p_0(x) + \omega(x)\, \hat{p}(x).
\end{equation}
A single upper bound $\omega$ is learned, and on a case-by-case basis the coefficient is reduced to the largest value that does not change the predicted class of the unaugmented input:
\begin{equation}
\omega(x) = \omega_{\text{pres}}(x) := \max\bigl\{\omega' \in [0, \omega] : \arg\max_c \tilde{p}_{\omega'}(x)_c = \arg\max_c p_0(x)_c \bigr\},
\label{eq:omega_pres}
\end{equation}
where $\tilde{p}_{\omega'}(x) = (1-\omega')\, p_0(x) + \omega'\, \hat{p}(x)$. Parameters $\boldsymbol{\theta}$ and $\omega$ are optimized on the validation set using NLL. The argmax-preserving reduction of Eq.~\eqref{eq:omega_pres} is applied both within the optimization loop and at inference.

\smallskip
\noindent\textbf{AugTTA}~\cite{shanmugam2021better} learns one non-negative weight $\theta_m \geq 0$ per individual augmentation (including the original at $m=0$), without grouping or $\omega$-blending:
\begin{equation}
\tilde{p}_{\text{AugTTA}}(x) = \text{normalize}\!\left(\sum_{m=0}^{M} \theta_m \, p_m(x)\right),
\end{equation}
where non-negativity is enforced via clamping and weights are optimized on the validation set using NLL.

\smallskip
\noindent\textbf{DualTTA} combines per-augmentation weighting (as in \emph{AugTTA}) with $\omega$-blending (as in \emph{V-ATTA}). These mechanisms address orthogonal failure modes: per-augmentation weights capture the unequal reliability of individual augmentations and severity levels, which type-level weights cannot, while $\omega$-blending explicitly anchors the output to the baseline prediction of the black-box model, which pure reweighting cannot. \emph{DualTTA} learns non-negative weights $\theta_m \geq 0$ for each augmentation $m = 1, \ldots, M$ (the baseline enters separately through blending):
\begin{equation}
\hat{p}(x) = \text{normalize}\!\left(\sum_{m=1}^{M} \theta_m \, p_m(x)\right),
\end{equation}
\begin{equation}
\tilde{p}_{\text{DualTTA}}(x) = \bigl(1 - \omega(x)\bigr)\, p_0(x) + \omega(x)\, \hat{p}(x),
\end{equation}
with the per-case blending coefficient
\begin{equation}
\omega(x) =
\begin{cases}
\omega, & \text{if } \max_c \hat{p}(x)_c - \max_c p_0(x)_c > \tau,\\[2pt]
\omega_{\text{pres}}(x), & \text{otherwise},
\end{cases}
\label{eq:dualtta_omega}
\end{equation}
where $\omega_{\text{pres}}(x)$ is the argmax-preserving reduction of Eq.~\eqref{eq:omega_pres} and $\tau = 0.10$. During weight optimization, only the second case is used, so that \emph{DualTTA} preserves the baseline argmax exactly as \emph{V-ATTA}. At inference, the first case additionally applies: when the aggregated prediction is substantially more confident than the baseline, the full learned $\omega$ is used without reduction, allowing the predicted class to change and correcting overconfident baseline errors. We fix $\tau = 0.10$ so that the predicted class changes only when the aggregated output is clearly more confident than the baseline, rather than from small fluctuations. This confidence-gated rule is present in neither parent method: \emph{V-ATTA} always preserves the baseline argmax, whereas \emph{AugTTA} never protects it. \emph{DualTTA} thus remains conservative by default while retaining the ability to correct overconfident baseline errors.

\subsection{Experimental Setup}

\paragraph{Datasets and Splits.}
We evaluate our framework on two binary classification tasks: pulmonary embolism (PE) detection using the INSPECT CTPA dataset \cite{huang2023inspect}, and intracranial hemorrhage (ICH) detection on the RSNA ICH Detection Challenge~\cite{rsna2019ich}. For ICH, we use 15{,}215/2{,}174/4{,}354 scans for train\-/vali\-dation\-/test, with 41\% disease prevalence. For PE, we use 12{,}716/1{,}838/3{,}676 scans with 24\% prevalence. All splits are done at the patient level. For ICH, we remove air and table via foreground cropping ($>-300$ HU), resample to $(0.8, 0.8, 1.0)\,\mathrm{mm}$ spacing using trilinear interpolation, and clip intensities to a subdural window $[-50, 150]$ HU. For PE, we resample to $(1.5, 1.5, 3.0)\,\mathrm{mm}$ spacing and clip to $[-1000, 1000]$ HU. All volumes are resized to $224 \times 224 \times 160$ voxels.

\paragraph{Black-Box Model Training.}
To simulate black-box classifiers, we fine-tune the image encoder of MERLIN, a CT foundation model~\cite{blankemeier2024merlin}, separately for PE and ICH using 3D augmentations: random flips, random 90-degree rotations, Gaussian noise, and intensity scaling. These augmentations are applied only during classifier training, not during evaluation-time TTA. For each task, we train three models with different random seeds from the same pretrained initialization. After training, we freeze all parameters and retain only the output probabilities for the validation and test sets. This setting simulates the scenario where institutions have access to labeled retrospective cases but cannot retrain proprietary models.
\noindent\paragraph{Aggregation Weight Training.} Learned aggregation methods operate directly on probability outputs from the frozen black-box classifier. Rather than training a complex model over augmented predictions, we learn only scalar augmentation weights, adding negligible overhead and requiring only a modest labeled validation set. Because logits are unavailable in the black-box setting, we adapt \emph{V-ATTA}~\cite{conde2023approaching} and \emph{AugTTA}~\cite{shanmugam2021better} to probability outputs. For \emph{V-ATTA}, \emph{AugTTA}, and \emph{DualTTA}, weights are
  optimized on the validation set using SGD (momentum 0.9, weight decay $10^{-4}$)
  with a learning rate of 0.01 for 50 epochs and batch size 500, minimizing NLL. GPS greedily selects up to 3 augmentations that minimize
  validation NLL.
  
\paragraph{Evaluation Protocol.}
We report mean and standard deviation of the Expected Calibration Error (ECE) with 10 equal-width bins spanning [0, 1] and Brier score over the three independently trained models. As reference baselines, we evaluate three calibration methods that, unlike our framework, require access to model internals or retraining: temperature scaling~\cite{guo2017calibration} (a single temperature fit per dataset on the validation set), MC dropout (dropout left active at inference, averaged over 100 stochastic forward passes), and deep ensembles (probabilities averaged across the three independently trained models). We further construct
  reliability diagrams using adaptive (quantile-based) binning, where bin edges
  are determined from each method's pooled predictions across seeds, such that
  each bin contains an approximately equal number of samples. We also ablate
  \emph{DualTTA} by training with a single augmentation type (all 3 severity
  levels) or a single severity level (all 6 types), comparing against the
  full 18-augmentation configuration to evaluate the contribution of
  augmentation diversity. 

\section{Results and Discussion}

\begin{table}[t]
    \centering
    \caption{Expected Calibration Error (ECE) and Brier Scores for PE and ICH across
different methods. Temperature scaling, MC dropout, and deep ensembles are included as non-black-box comparators. Values are mean ±std over three seeds.}
    \label{tab:ich_pe_results_id}
    \resizebox{\linewidth}{!}{
    \setlength{\tabcolsep}{7pt}
    \begin{tabular}{@{}lcccc@{}}
    \toprule
    & \multicolumn{2}{c}{\textbf{PE}} & \multicolumn{2}{c}{\textbf{ICH}} \\
    \cmidrule(lr){2-3}\cmidrule(lr){4-5}
    \textbf{Method}
    & \textbf{ECE} $\downarrow$ & \textbf{Brier} $\downarrow$
    & \textbf{ECE} $\downarrow$ & \textbf{Brier} $\downarrow$ \\
    \midrule

    No TTA
    & 0.239 $\pm$ 0.017 & 0.231 $\pm$ 0.005
    & 0.051 $\pm$ 0.025 & 0.081 $\pm$ 0.009 \\

    \midrule

    Temperature scaling
    & 0.242 $\pm$ 0.014 & 0.231 $\pm$ 0.006
    & 0.032 $\pm$ 0.005 & 0.076 $\pm$ 0.004 \\
    MC dropout
    & 0.240 $\pm$ 0.017 & 0.231 $\pm$ 0.005
    & 0.050 $\pm$ 0.025 & 0.080 $\pm$ 0.009 \\
    Deep ensembles
    & 0.224 $\pm$ 0.013 & 0.219 $\pm$ 0.007
    & 0.031 $\pm$ 0.012 & \textbf{0.069} $\pm$ \textbf{0.001} \\

    \midrule

    Equal TTA
    & 0.161 $\pm$ 0.043 & 0.198 $\pm$ 0.013
    & 0.058 $\pm$ 0.022 & 0.079 $\pm$ 0.007 \\
    Max TTA
    & 0.336 $\pm$ 0.027 & 0.293 $\pm$ 0.020
    & 0.229 $\pm$ 0.070 & 0.180 $\pm$ 0.033 \\
    GPS TTA
    & 0.206 $\pm$ 0.032 & 0.215 $\pm$ 0.012
    & 0.100 $\pm$ 0.032 & 0.097 $\pm$ 0.007 \\

    \midrule

    V-ATTA
    & 0.129 $\pm$ 0.030 & 0.193 $\pm$ 0.007
    & 0.030 $\pm$ 0.013 & 0.075 $\pm$ 0.004 \\
    AugTTA
    & 0.159 $\pm$ 0.041 & 0.197 $\pm$ 0.012
    & 0.043 $\pm$ 0.014 & 0.077 $\pm$ 0.004 \\
    DualTTA
    & \textbf{0.109} $\pm$ \textbf{0.041} & \textbf{0.185} $\pm$ \textbf{0.009}
    & \textbf{0.029} $\pm$ \textbf{0.011} & 0.074 $\pm$ 0.004 \\

    \bottomrule
    \end{tabular}
    }
\end{table}

Table~\ref{tab:ich_pe_results_id} compares calibration performance across the baseline model, TTA aggregation strategies, and non-TTA  comparator baselines. Among TTA methods, learned weighting methods (\emph{V-ATTA},
\emph{AugTTA}, and \emph{DualTTA}) consistently achieve the best ECE values across both tasks. \emph{DualTTA} and \emph{V-ATTA} yield the strongest improvements, reducing ECE
by 46–54\% relative to \emph{No TTA} on PE (0.239 → 0.109–0.129) and by 41–43\% on
ICH (0.051 → 0.029–0.030). Notably, \emph{DualTTA} achieves lower ECE than both of its constituent approaches on both tasks, supporting that per-augmentation weighting and baseline blending address complementary failure modes rather than being redundant.

Simpler aggregation strategies show more mixed behavior: \emph{Equal Weighting} provides modest gains, whereas \emph{Max Weighting} worsens calibration, increasing ECE to 0.336 on PE and 0.229 on ICH. This contrast shows that calibration gains are not an automatic consequence of augmentation, but depend on how augmented predictions are combined. Among non-TTA comparators,  Temperature scaling improves ICH calibration but provides little benefit for PE, while MC dropout performs similarly to the unmodified model. Deep ensembles improve both ECE and Brier score relative to \emph{No TTA}, achieving the lowest ICH Brier score.

\begin{figure}[t]
{\includegraphics[width=\linewidth]{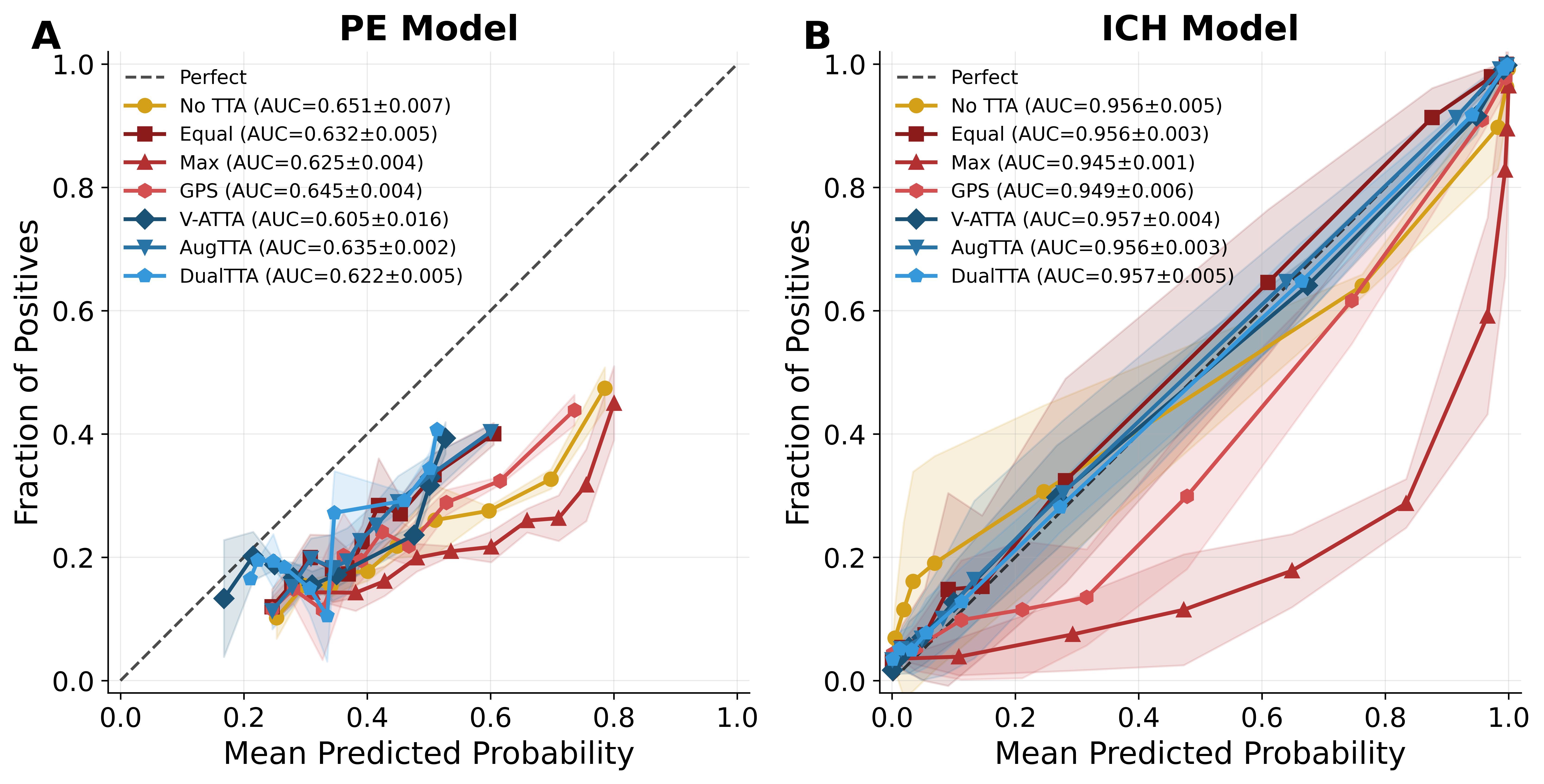}}
{\caption{Reliability diagrams for detecting pulmonary embolism (A) and intracranial hemorrhage (B) across TTA aggregation methods. Each
   curve shows the mean fraction of positive cases vs. the mean predicted probability
  across three model seeds. Shaded bands indicate the inter-seed
  variability (mean ± std across seeds). Adaptive (quantile-based) binning is used: bin edges are determined
   from each method's pooled predictions across seeds such that each bin
  contains an approximately equal number of samples, avoiding unreliable
  estimates in sparsely populated probability regions. The better the calibration, the closer to the diagonal.\label{fig:calibration_plots}}}
\end{figure}

The magnitude of calibration improvement differs across tasks. The PE
model exhibits a higher baseline calibration error (ECE=0.239), whereas the
ICH one is already relatively well-calibrated (ECE=0.051). Nonetheless, learned
aggregation reduces ECE on ICH, suggesting that the approach does not rely
on a large initial ECE to be effective. The small inter-seed standard deviations
indicate that these effects are stable and not driven by random initializations.
Fig.~\ref{fig:calibration_plots} provides a complementary view via reliability diagrams. On PE, all
methods deviate from the diagonal, reflecting task difficulty, but learned strategies (\emph{V-ATTA} and especially \emph{DualTTA}) produce a visibly closer alignment to
perfect calibration. On ICH, calibration curves are generally close to ideal across
methods, with \emph{Max Weighting} again demonstrating systematic overconfidence.

Aggregating predictions across augmentations does not systematically improve discrimination performance like it improves calibration. As shown in Fig.~\ref{fig:calibration_plots}, AUC values remain comparable across methods. On PE, learned aggregation methods achieve AUC similar to or slightly below baseline (e.g., \emph{AugTTA}: 0.635 vs.\ \emph{No TTA}: 0.651), while on ICH all methods cluster tightly around high baseline AUC ($\approx$0.94--0.96). This indicates that learned aggregation primarily adjusts confidence calibration rather than altering classification performance.

\subsection*{Ablation of Augmentation Types and Severity Levels}

\begin{figure}[t]
{\includegraphics[width=\linewidth]{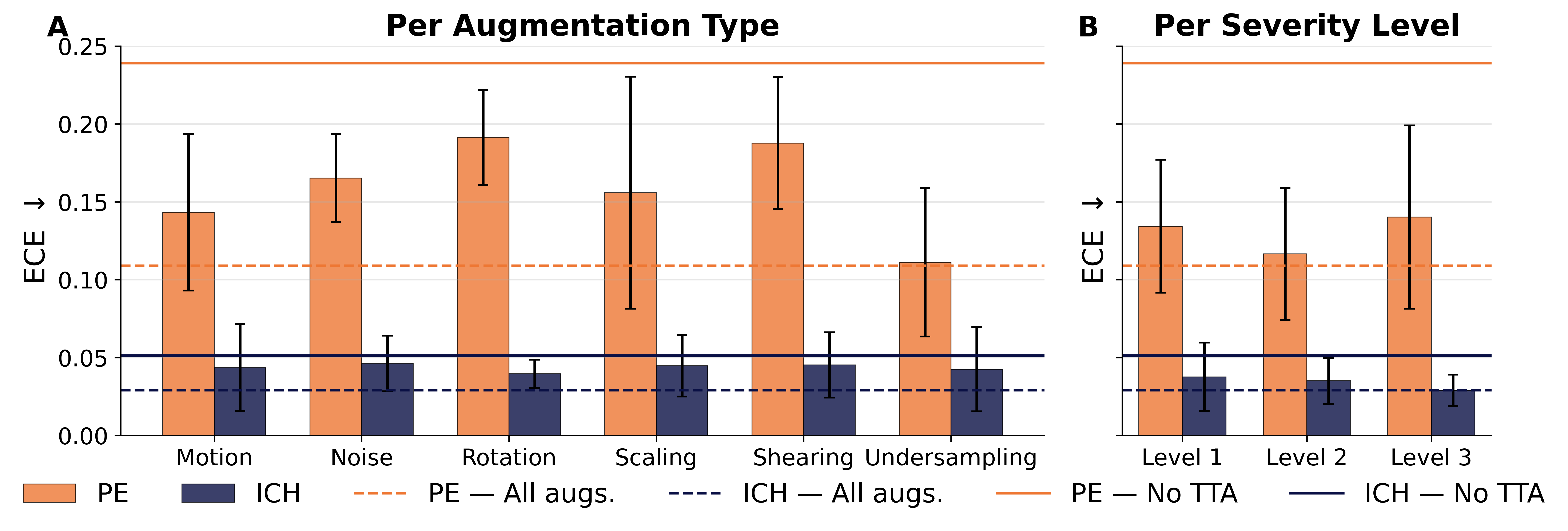}}
{\caption{\emph{DualTTA} ablation on ECE across augmentation types and severity levels for
  PE and ICH. (A) \emph{DualTTA} trained with a single augmentation type (all 3
  severity levels). (B) \emph{DualTTA} trained with all 6 augmentation types at a
  single severity level. Dashed lines indicate \emph{DualTTA} performance when trained
  with all 18 augmentations.
  Solid lines indicate ECE with \emph{No TTA}. Error bars show mean ± std across three seeds.
  Using all augmentations yields the best calibration, particularly for
  PE, while individual types and severity levels show comparable performance
  for ICH.\label{fig:ablation_plots}}}
\end{figure}

Fig.~\ref{fig:ablation_plots} examines how augmentation diversity contributes to calibration improvements. When \emph{DualTTA} is trained using only a single augmentation type, calibration performance on PE improves relative to \emph{No TTA} but does not match the full 18-augmentation configuration. A similar pattern is observed when training with all types at a single severity level. Combining geometric and acquisition-based perturbations across all severities yields the lowest ECE.

This pattern is more pronounced for PE than for ICH. For PE, the full augmentation set provides clear additional gains over individual types, suggesting complementary effects across perturbation types. For ICH, differences between augmentation subsets are smaller, consistent with the higher overall predictive performance. These findings indicate that calibration improvements are not attributable to a single augmentation type. Instead, combining diverse, clinically plausible perturbations and dynamically assigning each a relevant weight yields more stable probability estimates.

\section{Conclusion}

We presented a model-agnostic framework for improving calibration of radiology AI using clinically grounded test-time augmentation (TTA). A library of geometric and acquisition-based 3D CT perturbations was combined with learned aggregations, requiring no access to model internals. Across pulmonary embolism and intracranial hemorrhage detection, methods that learn per-augmentation weights (\emph{V-ATTA} and especially \emph{DualTTA}) consistently improved calibration, often outperforming standard uncertainty estimation techniques that require model internals. Ablation analysis confirmed that combining diverse perturbations yielded the most consistent improvements. However, our study is limited in that we evaluated simulated rather than true commercial black-box models. Despite this, our framework is applicable to real-world proprietary systems, and therefore supports the reliable interpretation of model outputs under black-box settings. Future work should evaluate performance under explicit cross-scanner and cross-institution shifts using real commercial systems.

\begin{credits}
\subsubsection{\ackname} This study was funded with research support from NIH grants R01 HL167974, R01HL169345, R01 AR077604, R01 EB002524, R01 AR079431, P41 EB 027060 and P50 HD118632.

\subsubsection{\discintname}
The authors have no competing interests to declare that are relevant to the content of this article.
\end{credits}

\bibliographystyle{splncs04}
\bibliography{miccai-bibliography}

\end{document}